# *Application of Multi-channel 3D-cube Successive Convolution Network for Convective Storm Nowcasting*

Wei Zhang
Ocean University of China
Qingdao, Shandong Province
China
weizhang@ouc.edu.cn

Lei Han
Ocean University of China
Qingdao, Shandong Province
China
hanlei@ouc.edu.cn

Juanzhen Sun
National Center for Atmospheric Research
Boulder, CO
USA
sunj@ucar.edu

Hanyang Guo
Ocean University of China
Qingdao, Shandong Province
China
ihcil@ouc.edu.cn

Jie Dai
Ocean University of China
Qingdao, Shandong Province
China
dayou@ouc.edu.cn

***Abstract*—very short-term weather forecasting or nowcasting has attracted substantial attention in various fields. Existing methods can nowcast storm advection based on radar data. Due to the limitations of the radar observations, it is still challenging to nowcast storm initiation and growth. However, as the real-time re-analysis meteorological data can now provide valuable atmospheric boundary layer thermal dynamic information, which is essential to predict storm initiation and growth. It is of great importance to leverage these re-analysis data.**

***This paper describes our first attempt to nowcast storm initiation, growth, and advection simultaneously under the framework of convolutional neural network using the very large multi-source meteorological data. To this end, we construct a multi-channel 3D-cube successive convolution network which leveraging both raw 3D radar and re-analysis data directly without any handcraft feature engineering. These data are formulated as multi-channel 3D cubes, to be fed into our network, which are convolved by cross-channel 3D convolutions. By stacking successive convolutional layers without pooling, we build an end-to-end trainable model for nowcasting. Experimental results show that deep learning methods achieve better performance than traditional extrapolation methods. The qualitative analyses of our approach show encouraging results of nowcasting of storm initiation, growth, and advection.***

***Keywords: Weather forecasting; Data Mining; Deep Learning; Convolutional Neural Network***

## I. Introduction

Very short-term convective storm forecasting, also known as nowcasting, focuses on time- and space-specific weather forecasts for periods of less than a few hours and has attracted substantial attention in various fields. Accurate nowcasts of convective storms have high impacts in many fields, such as agriculture, the aviation industry, and energy providers. Given the limitations of real-time convective-scale observation and the lack of a physical understanding of convective storm initiation and evolution, it remains a challenging problem [1, 2]. Recently, deep learning techniques have shown some potential in this area.

Existing nowcasting methods can be classified into two categories [1, 3]: numerical weather prediction (NWP) models and extrapolation techniques [4-12]. Although there has been great progress in NWP models, the method is still far from being adequate for nowcasting because of many problems, such as rapid model error growth at the convective scale [1, 13]. Extrapolation techniques primarily use radar data to extrapolate radar echoes to generate nowcasts. A typical example is the Thunderstorm Identification, Tracking, and Nowcasting (TITAN) algorithm [14], which has been used widely worldwide. TITAN uses 35 dBZ as the radar reflectivity threshold to identify, track, and nowcast storms. Generally, existing extrapolation techniques can forecast existing storm advection well. However, radar reflectivity observations alone do not provide sufficient atmospheric boundary layer (ABL) thermal dynamic information, making it difficult for extrapolation methods to predict convective storm initiation (CI) [1] and storm growth, which are important issues [1, 2].

With recent developments in meteorological data assimilation techniques [1, 15], real-time high-resolution numerical weather re-analysis systems can provide more accurate ABL thermal dynamic information, providing the opportunity to incorporate such data into storm initiation and growth nowcasting. Raw radar and re-analysis data are, in fact, now three-dimensional.

Weather forecasting is one of the original 'big data' problems. With large amounts of radar and re-analysis data, can we learn the patterns directly without handcraft feature engineering? Deep learning, especially convolutional neural networks (CNNs), has achieved successes with many challenging problems [16], and now makes it possible to achieve this.

Since AlexNet achieved great success in Imagenet classification [37], many researchers have further developed CNNs ([16], [35, 36, 39-42]). Szegedy et al. [38] developed a 22-layer CNN that achieves comparable classification results to the ImageNet benchmark. Another novel CNN is Network In Network (NIN) [34], which builds micro-networks with a multilayer perceptron to abstract the data within the receptive field. Some convolutional networks have been developed to recognize 3D rigid objects [20-23], such as chairs or desks, and use time as the third dimension ([17-19]) to develop their 3D convolutional networks for video analyses. Compared with the research objects above, convective storms have unique characteristics because they are non-rigid objects, they may initiate and dissipate in a short time, and they do not have a fixed shape. Thus, existing 3D convolutional networks are not directly applicable to our study.

[1] Here we use "convective initiation" or CI to refer to a storm that is grown from "scratch" (fewer than 35 dBZ radar echoes) rather than from an existing storm nearby with greater than 35 dBZ echoes.

In this study, we formulate the problem of nowcasting as a classification problem: that is, will a radar echo > 35 dBZ appear in a specific location in a short time? We present a multi-channel 3D-cube successive convolution network (3D-SCN) for convective storm nowcasting. Raw 3D radar and re-analysis meteorological data are used as multi-channel 3D cubes as our network input, which are convolved by cross-channel 3D convolutions. By stacking successive convolutional layers without pooling, we build an end-to-end trainable model for nowcasting.

## II. Related Work

Applications of deep learning in the field of atmospheric science are rare and those that have occurred are very recent, in the past 1–2 years. Liu et al. [24] used climate data to train a traditional CNN for detecting tropical cyclones and atmospheric rivers. In addition, Tao et al. [25] applied stacked denoising auto-encoders to reduce bias in satellite precipitation products and achieved significant improvements. Hossain et al. [26] used the same method to predict air temperature, based on the prior 24 h of hourly temperature data in northwestern Nevada. Finally, Grover et al. [27] proposed a hybrid method that combines DNN and a graphical model to forecast four weather variables (wind velocity, pressure, temperature, dew point) using balloon observation data.

Generally, the studies above focused on conventional weather forecast problems, applied to isolated weather stations, or detecting a synoptic weather phenomenon at hundreds-of-kilometers scale. These are not what the nowcasting community focuses on.

Shi et al. [3] and Klein et al. [28] took aim at the nowcasting problem using deep learning methods. Klein et al. [28] proposed a dynamic convolutional neural network that used four consecutive 2D radar reflectivity images to produce radar images of the next 10 min. Shi et al. [3] proposed a novel convolutional LSTM network, which also used consecutive 2D radar reflectivity images to produce several radar images for precipitation nowcasting.

The work of Shi et al. [3] and Grover et al. [27] can still be classified as extrapolation methods. With state-of-the-art techniques, they can nowcast storm advection very well using radar reflectivity data. Beyond storm advection forecasts, predictions of convective storm initiation (CI) and growth are other key issues for nowcasting. Owing to the limitations of radar observations, as mentioned previously, it remains very difficult for extrapolation methods to predict CI and storm growth [1, 2].

This study describes the first attempt to nowcast storm initiation, growth, and advection simultaneously under a deep learning framework using multi-source meteorological data directly, with no handcraft feature engineering.

## III. Preliminary Work

### A. Multi-source Meteorological Data

Raw radar reflectivity (*R*) data are actually three- dimensional (3D). However, most existing methods, such as those described previously [3, 28], only use 2D radar images, which may be a particular layer of 3D radar data, or a 2D projection of 3D radar data. Instead, in our study, we use raw 3D radar data.

In contrast to the use of radar data, very few previous studies have used real-time raw re-analysis meteorological data, which are often provided by state-of-the-art numerical weather re-analysis systems, such as VDRAS, proposed by the National Center for Atmospheric Research (NCAR) of America [1, 15]. Such re-analysis data include many retrieved meteorological variables, and each variable is also 3D. Compared with radar data, these data can provide valuable ABL thermal dynamic information for CI and storm growth nowcasting.

In this study, to predict storm initiation and growth as well as storm advection, beyond 3D raw radar reflectivity data (*R*), we also used two commonly used physical variables, *w* and *byc* of the VDRAS output in our study: *w* is the vertical velocity and *byc* is the buoyancy of an air parcel.

As a few statistics of *w* and *byc*, such as mean and standard deviation, do not provide as much information as the raw data itself, we used the raw 3D data of *w* and *byc* directly, with no handcraft feature engineering, so we could extract the maximum amount of useful information on convective weather. The characteristics of these multi-source 3D data needed to be considered carefully in the formulation of the nowcasting problem.

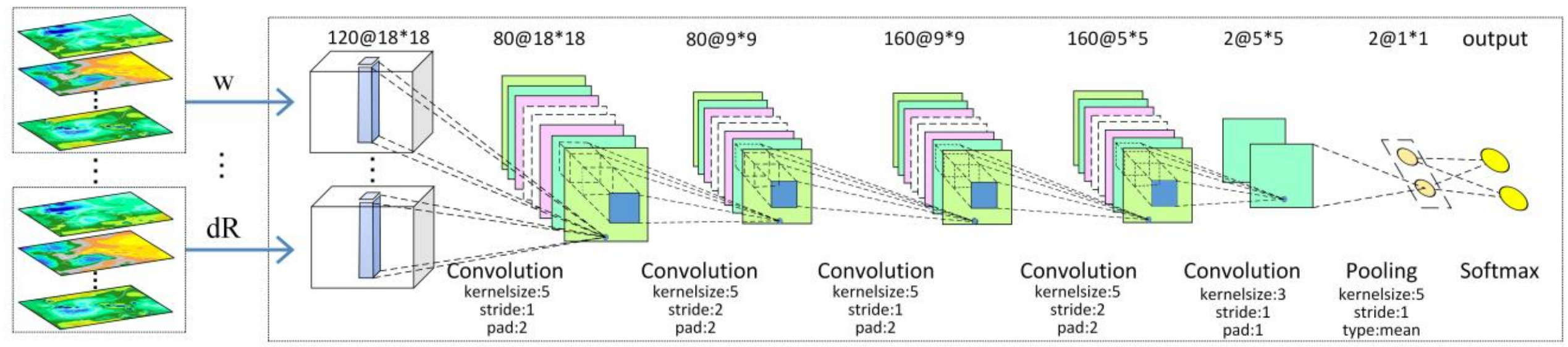

(w,dw,byc,dbyc,R,dR)

Figure 1: **Network Architecture of multi-channel 3D-SCN**

### *B. Formulation of the Nowcasting Problem*

The goal of nowcasting is to use real-time multi-source meteorological data to make time- and space-specific forecasts for periods of less than a few hours; here, we focus on a 30-min forecast which is commonly used for nowcasting test [14]. The study domain in this paper is in the Colorado Front Range area, USA, which is considered as a $31\times39$ grid (31 rows, 39 columns). Each cell in the grid has 36 pixels ($6\times6$ km). Now, we describe how to construct 3D cubes as input.

**3D Cubes** Because weather phenomena are continuous in space – each cell is impacted by its neighboring cells – we take a window $18\times18$ km, centered in a cell, to take such spatial impacts into account. We sought to use raw 3D radar and re-analysis data as input. For each $18\times18$ km window, we needed to use all 20 slices of data to include information at different altitudes. Each $18\times18\times20$ data volume of *w*, *byc*, and *R* is referred to as a 3D cube.

The temporal trend contains information about storm development and thus can also play an important role in convective nowcasting. For convective weather, the temporal trends of two consecutive datasets can fulfill our needs [2]. Thus, we also used the temporal trend of *w*, *byc*, and *R* as input into the deep network. For example, for two consecutive 3D cubes of *w*, we can get *dw* through the point-to-point difference:

$$dw = w_t - w_{t-15\min}$$

Here, the temporal interval of two consecutive *w* is 15 min. Finally, the data sample *X* is a six-channel signal:

$$X = (x_1, x_2, x_3, x_4, x_5, x_6) = (w, dw, byc, dbyc, R, dR) \quad (1)$$

**Class Label** For a cell at time t, if there is a radar echo > 35 dBZ at time t + 30 min, the 3D cube in the window centered in this cell is labeled “1,” meaning “a convective storm will happen in this cell in 30 min.” Otherwise, this cell will be labeled “0,” meaning “no convective storm will occur in this cell in 30 min.” Finally, the nowcasting problem can be transformed into a classification problem: will a radar echo > 35 dBZ appear in a cell within 30 min?

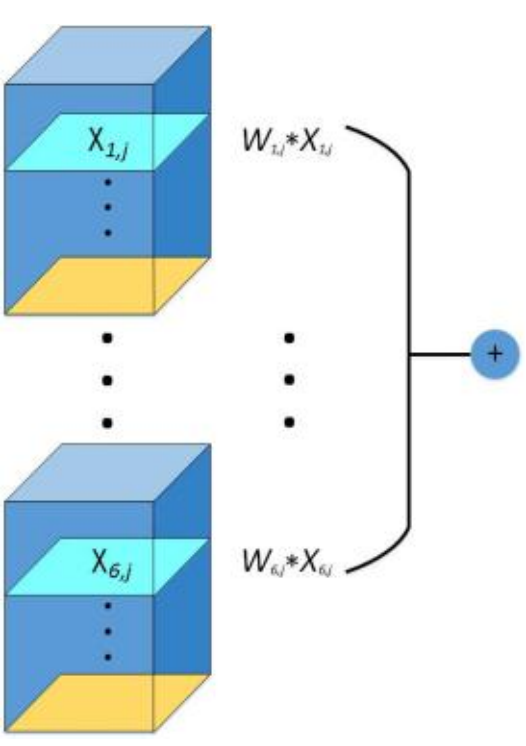

Figure 2: **Convolution on six channel 3D cubes**

## IV. The Model

### *A. Network Architecture*

Figure 1 shows our network architecture. The proposed 3D-SCN has five successive convolutional layers with no pooling between two adjacent convolutional layers. There is only one global average pooling on top of the last feature map. The kernel size is $5\times5$ while $3\times3$ was chosen for the last convolutional layer. We set stride as 1 for odd-numbered convolutional layers and 2 for even-numbered layers. The stochastic gradient descent (SGD) algorithm was used for learning with a learning rate of 0.001.

### *B. Cross-Channel 3D Convolution*

**3D Cubes as Input** Note that each data point *X* is a six-channel signal:

$$X = (x_1, x_2, x_3, x_4, x_5, x_6) = (w, dw, byc, dbyc, R, dR)$$

Each feature $x_i$ is a 3D cube ($18\times18\times20$) with 20 2D slices. The projection of $x_i$ is an $18\times18$-km window.

These six coupled physical variables (or channels) play different roles in the convective weather process. Each is important, but does not determine the weather by itself. All channels act together to determine the weather. Thus, a suitable convolution strategy to convolve different channels is essential. This is fulfilled by cross-channel 3D convolution, as described below.

**First Layer** For the first convolutional layer, there are 80 kernels. Here, we can establish the following equation:

$$X_k^1 = \text{ReLU}(\sum_j \sum_i W_{ijk}^1 * X_{ij} + b_k^1), k = 1, ..., 80 \quad (2)$$

where i is the index of channel, j is the index of slice within one channel, and k is the index of feature maps. $b_k^1$ is the linear bias, $X_k^1$ is the result feature map, $W_{ijk}^1$ is the weight matrix, and $X_{ij}$ is the input 2D slice of layer j within channel i. An illustration of $X_{ij}$ and $W_{ijk}^1$ is shown in Figure 2.

Because the altitude information for each meteorological variable is important for convective weather, for a fixed altitude, we convolved 2D slices on different channels to generate a layer-feature map. Thus, we can obtain 20 layer-feature maps. Then, we convolve all 20 layer-feature maps to provide an overall feature map. In this way, we accomplish cross-channel 3D convolution.

**Other Layers** Unlike the first layer, the convolution of other layers is performed as described previously [29], as follows:

$$X_q^l = \text{ReLU}(\sum_p W_{pq}^l * X_p^{l-1} + b_q^l) \quad (3)$$

where $W_{pq}^l$ is the weight matrix, p is the kernel in layer $l$-1, q is the kernel in layer $l$, and $X_p^{l-1}$ is the feature map. Equation (2) represents 3D convolution, whereas equation (3) represents 2D convolution.

### C. *The Fused Convolution*

In a previous paper [30], Springenberg et al. pointed out that pooling can simply be replaced by a convolutional layer with an increased stride with no loss in accuracy. References [31] and [32] indicate that any N×N convolution followed by a pooling operation can be fused into a single (N+1)×(N+1) convolution with stride set to 2, which could avoid about half of the computational cost of CNN.

In our case, we use the fused 5×5 convolution operation [stride = 2 with the convolution equation (3)], which saved ~63% of the computational cost versus conventional 4×4 convolution and pooling. Thus, the overall computational cost of successive convolution is less than with traditional CNN and NIN.

## V. Results and Discussion

### A. *Datasets*

Figure 3 shows our study domain. The data used in this study included radar reflectivity data from the KFTG WSR-88D radar located in Denver, USA, and VDRAS re-analysis data. The data for seven historic heavy rainfall events in the Colorado front range area used in this study (8–9 August, 2008; 28–29 July, 2010; 9–10 August, 2010; 13–14 July, 2011; 14–15 July, 2011; 6–7 June, 2012; and 7–8 July, 2012) were collected from a retrospective study of historical heavy rain/flash flood cases conducted by the Short Term Explicit Prediction (STEP)[2] Program of NCAR. The radar reflectivity images of the event on July 7, 2012, are shown in Figure 4 with a 30-min interval to show the evolution of a convective system. It can be seen that it is quite challenging to forecast storm initiation and growth. These radar images were also used as the ground truth to confirm the forecast.

It should be noted that although this study is performed over Denver area, it is also applicable to other places.

### B. *Nowcasting Metrics*

We used the contingency table approach [14] to evaluate the short-term forecasts, as is used commonly in weather forecasting. The probability of detection (POD), false alarm ratio (FAR), and critical success index (CSI) were calculated. POD, FAR, and CSI are similar to precision and recall, which are two skill scores used in the machine-learning field. In a previous paper [33], it was pointed out that, for low-frequency events, such as severe weather warnings, CSI is the better choice.

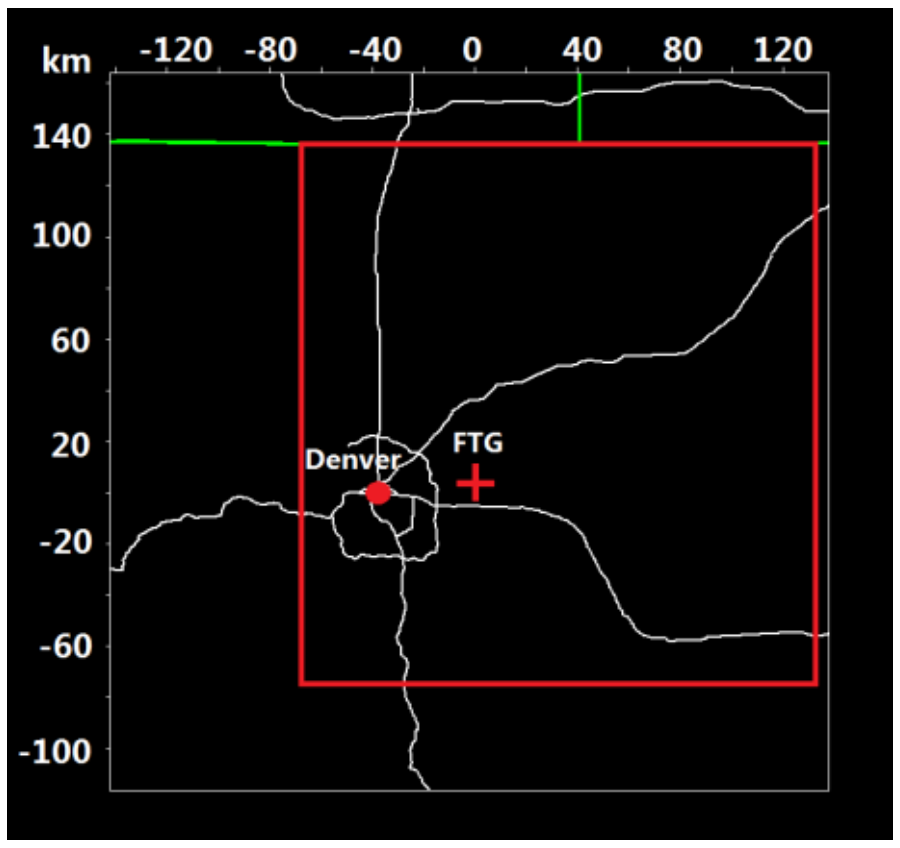

Figure 3: **Our study domain** (red rectangle). The location of the KFTG radar near Denver, USA, is shown by the red cross. The green lines indicate the state borders and the white lines are the major highways.

POD, FAR, and CSI are defined as:
POD = hits / (hits + misses) = recall
FAR = false alarms / (hits + false alarms) = 1 − precision
CSI = hits / (hits + misses + false alarms)

Here, in each cell, a hit occurs when this cell is classified as 1 (active) and there is radar echo greater than 35 dBZ in 30 min in the same cell (active), a miss occurs when the truth cell is active while the forecast cell is inactive, and a false alarm occurs when the truth cell is inactive while the forecast cell is active.

[2] http://www.rap.ucar.edu/projects/step/

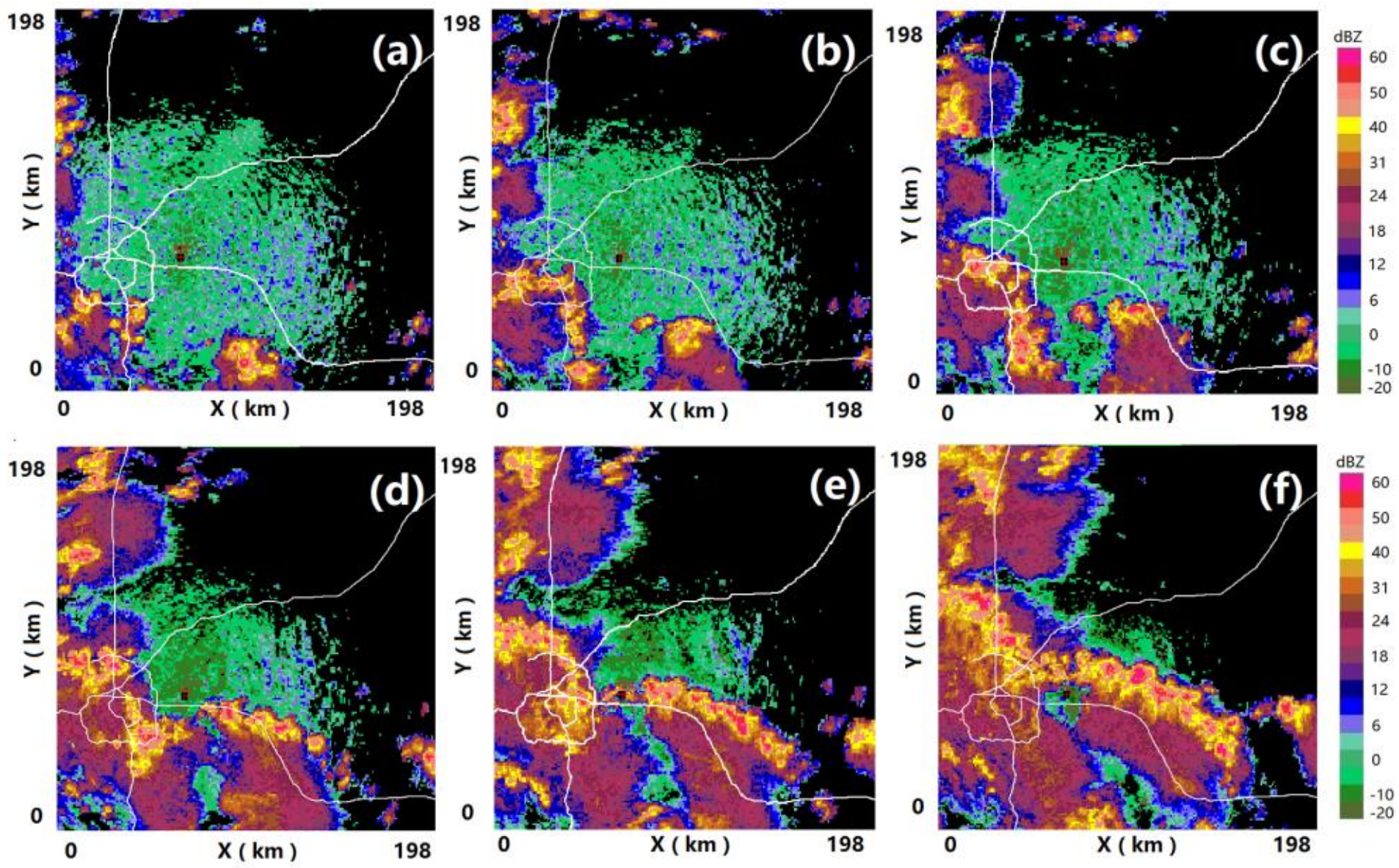

Figure 4: **Evolution of a convective weather system.** The radar reflectivity images are shown at (a) 21:30, (b) 22:00, (c) 22:30, (d) 23:00, (e) 23:30 and (f) 24:00 on 7 July 2012.

### C. *Five-fold Cross-Validation*

| | CSI(±σ) | POD(±σ) | FAR(±σ) |
|---|---|---|---|
| 3D-NIN | 0.443(±0.0055) | 0.704(±0.0197) | 0.455(±0.0192) |
| 3D-SCN | 0.443(±0.0026) | 0.678(±0.0136) | 0.438(±0.0091) |

Table 1: **Comparison of means and standard deviations**

| Date | POD | | FAR | | CSI | |
|---|---|---|---|---|---|---|
| | 3D-SCN | TITAN | 3D-SCN | TITAN | 3D-SCN | TITAN |
| 20120606 | 0.63 | 0.52 | 0.49 | 0.46 | 0.39 | 0.36 |
| 20120707 | 0.68 | 0.54 | 0.37 | 0.37 | 0.48 | 0.41 |

Table 3: **Verification statistics** for the 3D-SCN and TITAN nowcast for the cases of 6 June and 7 July 2012

We compare two methods in this section: 3D-NIN and 3D-SCN. We constructed 3D-NIN using our **cross-channel 3D convolution in the first convolutional** layer of NIN. The other parts of NIN are similar to NIN-Imagenet[3].

We randomized the sample set by shuffling all samples for cross-validation without considering whether they belonged to the same time. All 322,813 instances were subjected to 5-fold cross-validation (5-CV) training and tests.

The means and standard deviations (σ) for CSI, POD, and FAR are shown in Table 2. 3D-SCN and 3D-NIN achieved the same CSI value, but 3D-SCN converged faster. We ran a total of 100,000 iterations. For every 1,000 iterations, we computed CSI, POD, and FAR on the test set. Table 1 shows the iterations and training time where CSI reached its 'best' value. The average training time for 3D-SCN was much less than for 3D-NIN. This is not surprising because the number of 3D-NIN parameters was about 2.24 M, whereas 3D-SCN has only 1.36 M parameters.

Furthermore, 3D-SCN was more stable than 3D-NIN. It can be seen in Table 1 that the training time of 3D-SCN was ~3–4,000s, whereas that of 3D-NIN was ~5–15,000s. Moreover, 3D-SCN had a smaller standard deviation (Table 2).

| 3D-NIN | | | | | | |
|---|---|---|---|---|---|---|
| 5-CV | 1 | 2 | 3 | 4 | 5 | Average |
| Iterations | 66000 | 85000 | 32000 | 42000 | 42000 | |
| Time(seconds) | 12078 | 15385 | 5824 | 7644 | 7686 | 9723 |
| 3D-SCN | | | | | | |
| Iterations | 34000 | 25000 | 25000 | 28000 | 28000 | |
| Time(seconds) | 4325 | 3175 | 3175 | 3556 | 3556 | 3557 |

Table 2: **Iteration numbers and training time** of 3D-NIN and 3D-SCN

### D. *Consecutive Prediction Test*

We trained the 3D-SCN using 257,521 samples collected from five historic heavy rainfall events (8–9 August, 2008; 28–29 July, 2010; 9–10 August, 2010; 13–14 July, 2011; 14–15 July, 2011). Then, we used the trained models to make consecutive 30-min predictions for another two events (6–7 June, 2012; 7–8 July, 2012). Because this method uses another year's data for testing, it is more difficult than the cross-validation. However, this is the most often used verification method in the field of atmospheric sciences and is more practical.

[3] https://github.com/BVLC/caffe/wiki/Model-Zoo

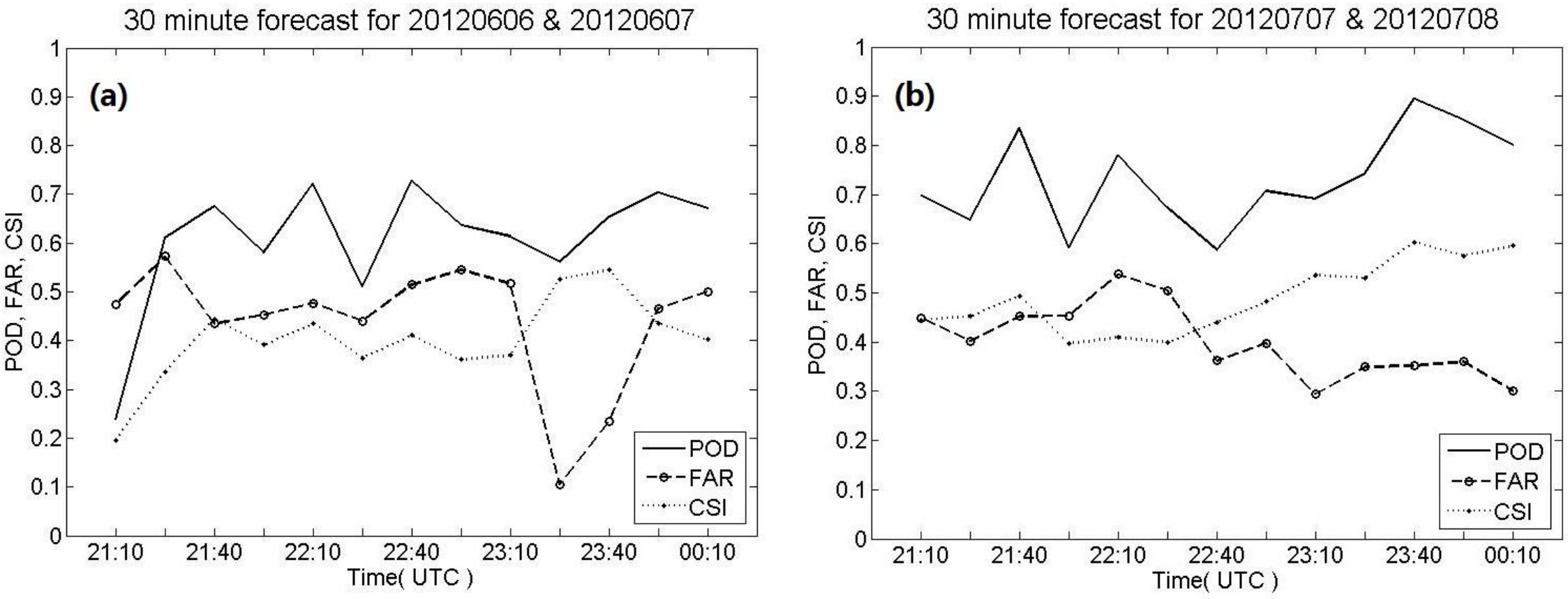

Figure 5: **Nowcasting statistics** for 6-7 June (a) and 7-8 July (b) 2012

**Quantitative Analysis** The time series statistics for CSI, POD, and FAR for the event on 6–7 June, 2012, are shown in Figure 5(a). Note that the POD and CSI values increased with time, partially because the storms grew larger. At the storm initiation stage (21:00–22:00 UTC), the POD and CSI values were relatively low. When storm grew larger with time, the POD and CSI values increased. This is reasonable because larger convective systems are relatively easier to forecast. Similar results were found for the event on 7–8 July, 2012.

Table 3 shows the verification statistics for the 3D-SCN and TITAN nowcast. 3D-SCN achieved higher POD and similar FAR than TITAN, leading to better CSI value which means 3D-SCN had better overall performance.

Since the receiver operating characteristic (ROC) curve is insensitive to the prior distribution of classes, it is useful to evaluate class-imbalance classification [43-44], such as nowcasting problem (in our case the positive samples are only about 5% of all samples). Figure 6 shows the ROC curves of the cases of 6 June and 7 July 2012. The area under ROC curve (AUC) of each case is greater than 0.9, which means 3D-SCN is highly predictive [45-46].

**Qualitative Analysis** Here, we present the 30-min forecast results for storm advection, initiation, and growth.

**1) Storm Advection**. Figure 7 shows consecutive radar images overlaid with 3D-SCN 30-min forecasts (red cells) and the corresponding verifications. Black cells represent the correctly predicted cells. Comparing Figures 6(a) and (d) at 22:55, we can see that the 3D-SCN forecasts captured the advection of line storms very well (indicated by the white arrows). The results at 23:10 and 23:25 show that the 3D-SCN forecasts continue to capture the storm movement quite well.

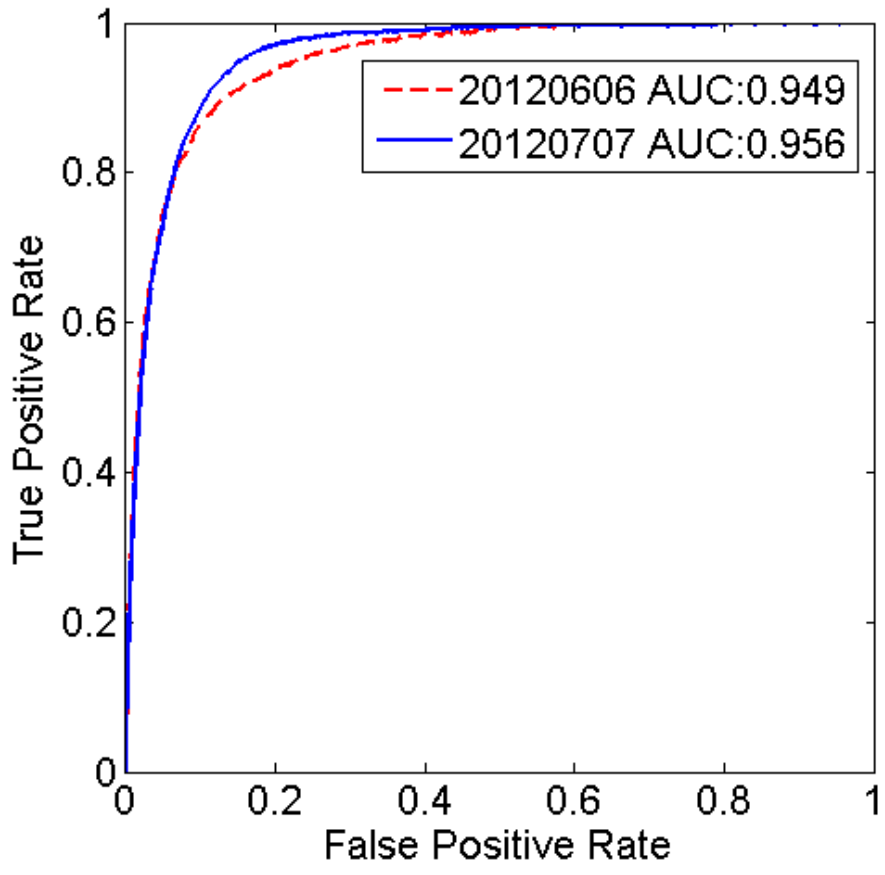

Figure 6: The **receiver operating characteristic** (ROC) of 3D-SCN prediction for the cases of 6 June and 7 July 2012

**2) Storm Initiation and Growth**. Figure 8 shows examples of convective initiation (CI). The black cells in the right panel show good agreement between the 3D-SCN forecast and the observed storm radar echoes. Figure 9 shows a case of storm growth. This storm grew significantly within 30 min, making it very hard to predict using extrapolation methods. The black cells in Figure 9b confirm that the 30-min 3D-SCN forecast agreed well with the storm radar echoes.

This is encouraging because the use of the real-time 3D re-analysis meteorological variables as inputs into our deep network made it possible to forecast storm initiation and growth well, while it is difficult for extrapolation methods to forecast CI and growth [2].

accuracy will remain challenging in future work.

## VI. Conclusions and Future Work

This study was aimed at nowcasting storm initiation and growth as well as storm advection under a deep learning framework. We present a multi-channel 3D-cube successive convolution network (3D-SCN) for convective storm nowcasting. We formulated the problem of nowcasting as a classification problem: that is, will a radar echo > 35 dBZ appear in a cell (6×6 km) in the short term? Raw 3D radar and re-analysis meteorological data were used as our network input. Cross-channel 3D convolution was used to convolve these multi-source data. By stacking successive convolutional layers without pooling, we built an end-to-end trainable model for nowcasting.

We compared three methods in our experiments. The results showed that both deep learning methods (3D-SCN and 3D-NIN) achieved better performance than a traditional extrapolation method. 3D-SCN showed comparably favorable performance with 3D-NIN, but 3D-SCN was more stable in training and required a much shorter training time. The qualitative analyses of 3D-SCN showed encouraging results of nowcasting of storm initiation, growth, and advection.

Although 3D-SCN can predict CI, its success is still limited. Because the duration of CI is very short compared with the whole lifetime of a storm, and its area is very small, it is difficult to collect enough training data for CI cases. Improving CI nowcast accuracy will remain challenging in future work.

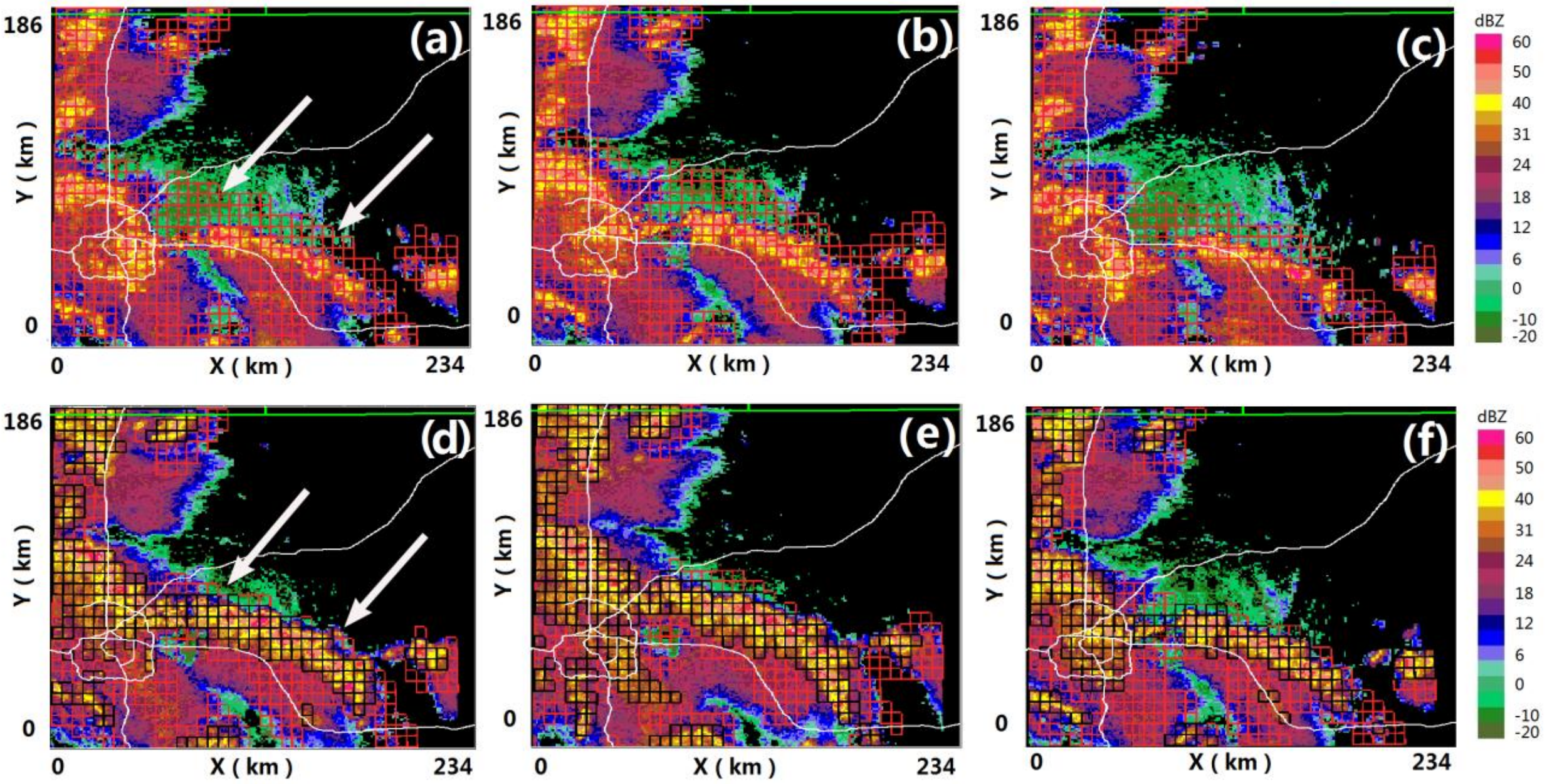

Figure 7**: Radar image and forecast results.** The red cells represent the 30-min forecasts. The cells at bottom are marked in black to represent those being correctly predicted. (a), (b) and (c) are the 30-min forecasts at issue time, 22:55, 23:10 and 23:25 respectively. (d), (e) and (f) are these same forecasts super-positioned over radar image at verification time 23:25, 23:40 and 23:55 respectively.

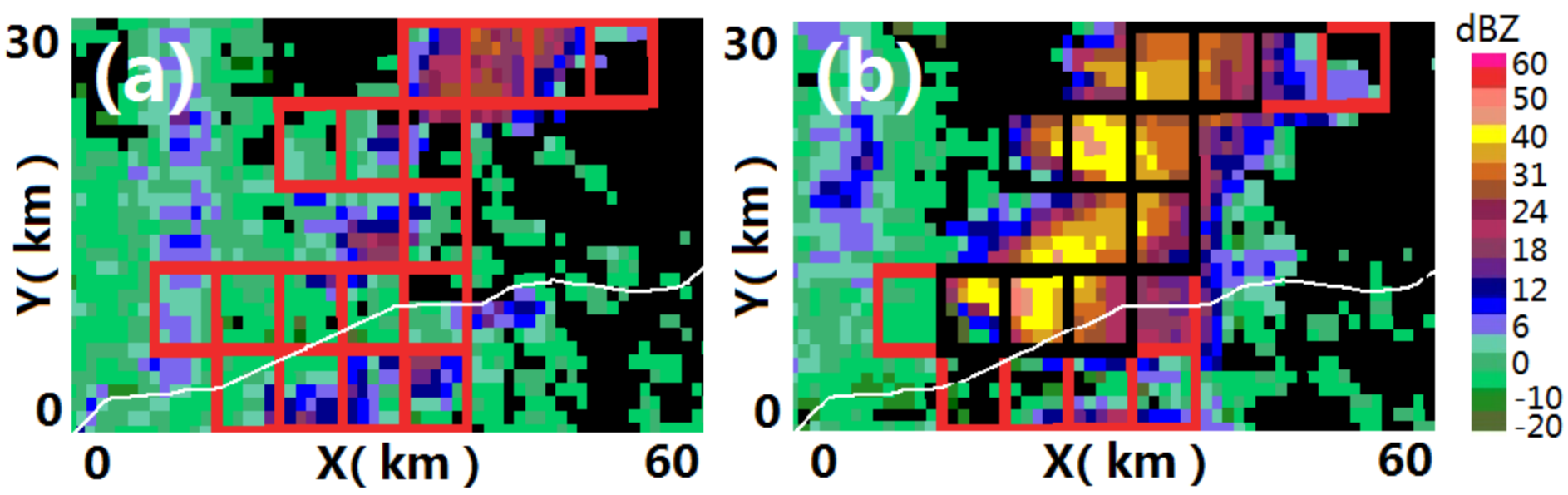

Figure 8: Same as Figure 6 but over a different sub-domain at the 30-min nowcast issue time (a) and at verification time (b) to show an example of CI nowcast.

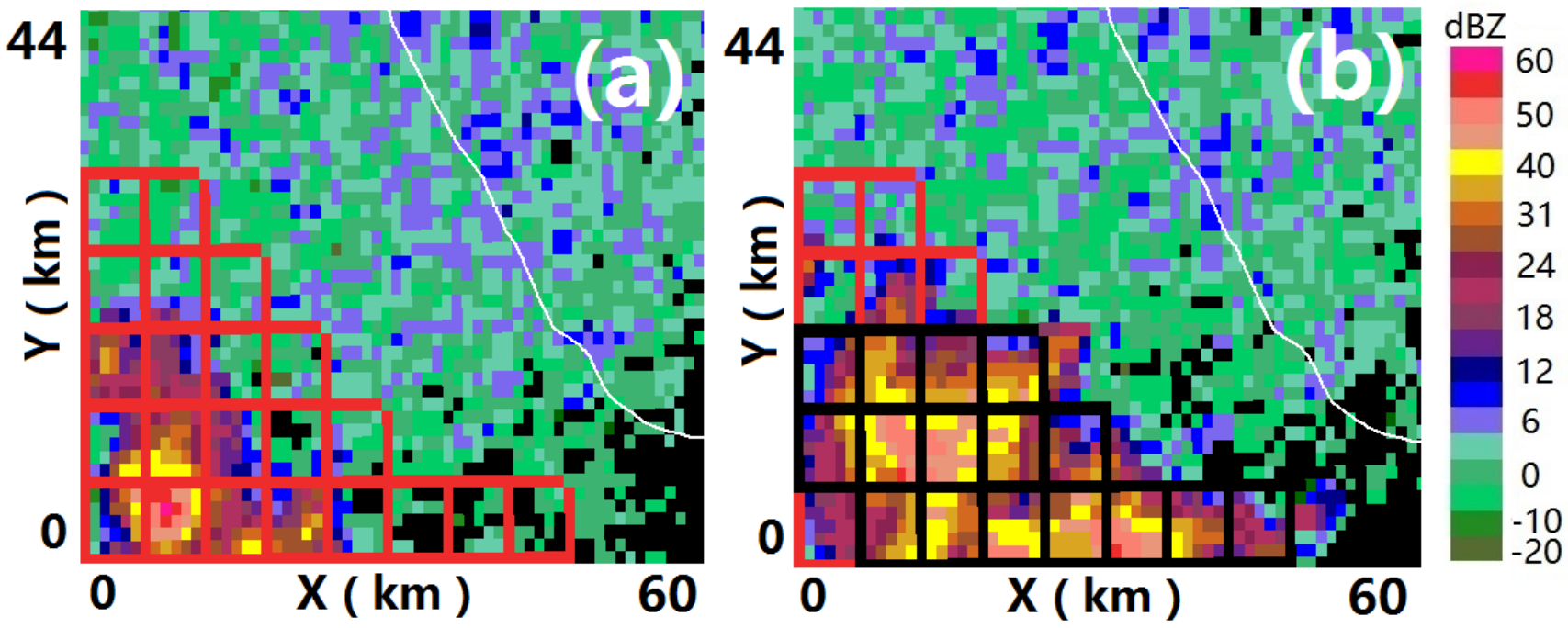

Figure 9: Same as Figure 6 but over a different sub-domain at the 30-min nowcast issue time (a) and at verification time (b) to show an example of storm growth nowcast.

## References

[1] J. Sun, M. Xue, J. W. Wilson, I. Zawadzki, S. P. Ballard, J. Onvlee-Hooimeyer, P. Joe, D. M. Barker, P. W. Li, B. Golding, M. Xu, and J. Pinto, "Use of nwp for nowcasting convective precipitation: Recent progress and challenges," Bull. Am. Meteorol. Soc., vol. 95, no. 3, pp. 409–426, 2014.

[2] J. W. Wilson, N. A. Crook, C. K. Mueller, J. Sun, and M. Dixon, "Nowcasting Thunderstorms: A Status Report," Bull. Am. Meteorol. Soc., vol. 79, no. 10, pp. 2079–2099, 1998.

[3] X. Shi, Z. Chen, and H. Wang, "Convolutional LSTM Network : A Machine Learning Approach for Precipitation Nowcasting," arXiv, pp. 1–11, 2015.

[4] R. E. Rinehart and E. T. Garvey, "Three-dimensional storm motion detection by conventional weather radar," Nature, vol. 273. pp. 287–289, 1978.

[5] J. D. Tuttle and G. B. Foote, "Determination of the boundary layer airflow from a single Doppler radar," J. ATMOSPHERIC & OCEANIC TECHNOLOGY, vol. 7, no. 2, NaN, 1990. pp. 218–232, 1990.

[6] L. Li, W. Schmid, and J. Joss, "Nowcasting of Motion and Growth of Precipitation with Radar over a Complex Orography," Journal of Applied Meteorology, vol. 34, no. 6. pp. 1286–1300, 1995.

[7] E. S. T. Lai, "Reprint 324 TREC Application in Tropical Cyclone Observation," no. August 1997, pp. 1–7, 1999.

[8] C. L. Bjerkaas and D. E. Forsyth, "Real-time automated tracking of severe thunderstorms using Doppler weather radar," in BULLETIN OF THE AMERICAN METEOROLOGICAL SOCIETY, 1979, vol. 60, no. 5, p. 533.

[9] G. L. Austin and A. Bellon, "Very-short-range forecasting of precipitation by the objective extrapolation of radar and satellite data," Nowcasting, pp. 177–190, 1982.

[10] J. Handwerker, "Cell tracking with TRACE3D - A new algorithm," Atmos. Res., vol. 61, no. 1, pp. 15–34, 2002.

[11] L. Han, S. Fu, L. Zhao, Y. Zheng, H. Wang, and Y. Lin, "3D convective storm identification, tracking, and forecasting - An enhanced TITAN algorithm," J. Atmos. Ocean. Technol., vol. 26, no. 4, pp. 719–732, 2009.

[12] D. Rosenfeld, "Objective method for analysis and tracking of convective cells as seen by radar," J. Atmos. Ocean. Technol., vol. 4, no. September, 1987.

[13] M. L. Weisman, C. Davis, W. Wang, K. W. Manning, and J. B. Klemp, "Experiences with 0–36-h Explicit Convective Forecasts with the WRF-ARW Model," Weather Forecast., vol. 23, no. 3, pp. 407–437, 2008.

[14] M. Dixon, G. Wiener, M. Dixon, and G. Wiener, "TITAN: Thunderstorm Identification, Tracking, Analysis, and Nowcasting—A Radar-based Methodology," J. Atmos. Ocean. Technol., vol. 10, no. 6, pp. 785–797, 1993.

[15] [J. Sun, M. Chen, and Y. Wang, “A Frequent-Updating Analysis System Based on Radar, Surface, and Mesoscale Model Data for the Beijing 2008 Forecast Demonstration Project,” Weather Forecast., vol. 25, no. 6, pp. 1715–1735, 2010.
[16] Y. Lecun, Y. Bengio, and G. Hinton, “Deep learning,” Nature, vol. 521, no. 1, pp. 436–444, 2015.
[17] [S. Ji, M. Yang, and K. Yu, “3D Convolutional Neural Networks for Human Action Recognition.,” Pami, vol. 35, no. 1, pp. 221–31, 2013.
[18] D. Tran, L. Bourdev, R. Fergus, L. Torresani, and M. Paluri, “Learning Spatialtemporal Features with 3D Convolutional Networks,” IEEE Int. Conf. Comput. Vis., 2015.
[19] A. Karpathy, G. Toderici, S. Shetty, T. Leung, R. Sukthankar, and L. Fei-Fei, “Large-Scale Video Classification with Convolutional Neural Networks,” 2014 IEEE Conference on Computer Vision and Pattern Recognition (CVPR). pp. 1725–1732, 2014.
[20] Z. Wu, S. Song, A. Khosla, F. Yu, L. Zhang, X. Tang, and J. Xiao, “3D ShapeNets: A deep representation for volumetric shapes,” Proc. IEEE Comput. Soc. Conf. Comput. Vis. Pattern Recognit., vol. 07–12–June, pp. 1912–1920, 2015.
[21] S. Song and J. Xiao, “Deep Sliding Shapes for Amodal 3D Object Detection in RGB-D Images,” arXiv Prepr., pp. 1–8, 2015.
[22] D. Maturana and S. Scherer, “VoxNet: A 3D Convolutional Neural Network for Real-Time Object Recognition,” pp. 922–928, 2015.
[23] B. Leng, Y. Liu, K. Yu, X. Zhang, and Z. Xiong, “3D object understanding with 3D Convolutional Neural Networks,” Inf. Sci. (Ny)., vol. 366, pp. 188–201, 2016.
[24] W. C. Yunjie Liu, Evan Racah, Prabhat, Joaquin Correa, Amir Khosrowshahi, David Lavers, Kenneth Kunkel, Michael Wehner, “Application of Deep Convolutional Neural Networks for Detecting Extreme Weather in Climate Datasets,” 2016.
[25] Y. Tao, X. Gao, K. Hsu, S. Sorooshian, and A. Ihler, “A Deep Neural Network Modeling Framework to Reduce Bias in Satellite Precipitation Products,” J. Hydrometeorol., vol. 17, no. 3, pp. 931–945, 2016.
[26] M. Hossain, B. Rekabdar, S. J. Louis, and S. Dascalu, “Forecasting the weather of Nevada: A deep learning approach,” in Proceedings of the International Joint Conference on Neural Networks, 2015, vol. 2015–Septe, pp. 2–7.
[27] A. Grover and E. Horvitz, “A Deep Hybrid Model for Weather Forecasting,” Proc. 21th ACM SIGKDD Int. Conf. Knowl. Discov. Data Min., pp. 379–386, 2015.
[28] B. Klein, L. Wolf, and Y. Afek, “A Dynamic Convolutional Layer for short rangeweather prediction,” Proc. IEEE Comput. Soc. Conf. Comput. Vis. Pattern Recognit., vol. 07–12–June, pp. 4840–4848, 2015.
[29] J. Bouvrie, “Notes on convolutional neural networks,” In Pract., pp. 47–60, 2006.
[30] J. T. Springenberg, A. Dosovitskiy, T. Brox, and M. Riedmiller, “Striving for Simplicity: The All Convolutional Net,” Iclr, pp. 1–14, 2015.
[31] F. Mamalet, S. Roux, and C. Garcia, “Real-time video convolutional face finder on embedded platforms,” Eurasip J. Embed. Syst., vol. 2007, 2007.
[32] F. Mamalet, S. Roux, and C. Garcia, “Embedded facial image processing with Convolutional Neural Networks,” Proc. 2010 IEEE Int. Symp. Circuits Syst., pp. 261–264, 2010.
[33] J. T. Schaefer, “The critical success index as an indicator of warning skill,” Weather Forecast., vol. 5, no. 4, pp. 570–575, 1990.
[34] M. Lin, Q. Chen, and S. Yan, “Network In Network,” arXiv Prepr., p. 10, 2013.
[35] Le Cun, B. Boser, et al. "Handwritten digit recognition with a back-propagation network."Advances in neural information processing systems. 1990.
[36] LeCun, Yann, et al. "Gradient-based learning applied to document recognition." Proceedings of the IEEE 86.11 (1998): 2278-2324.
[37] A. Krizhevsky, I. Sutskever, and G. E. Hinton. "Imagenet classification with deep convolutional neural networks". In Advances in Neural Information Processing Systems (NIPS), pages 1097–1105, 2012.
[38] C. Szegedy, W. Liu, Y. Jia, et al."Going deeper with convolutions." Proceedings of the IEEE Conference on Computer Vision and Pattern Recognition. 2015.
[39] K. Simonyan and A. Zisserman. Very deep convolu- tional networks for large-scale image recognition. In Internaltional Conference on Learning Representation (ICLR), 2015
[40] P. Sermanet, D. Eigen, X. Zhang, M. Mathieu, R. Fer- gus, and Y. LeCun. Overfeat: Integrated recognition, localization and detection using convolutional networks. In International Conference on Learning Representa- tions (ICLR), 2014
[41] R. Girshick, J. Donahue, T. Darrell, and J. Malik. Rich feature hierarchies for accurate object detection and se- mantic segmentation. In Proceedings of the IEEE Con- ference on Computer Vision and Pattern Recognition (CVPR), pages 580–587, 2014
[42] R. Girshick, J. Donahue, T. Darrell, and J. Malik. Rich feature hierarchies for accurate object detection and se- mantic segmentation. In Proceedings of the IEEE Con- ference on Computer Vision and Pattern Recognition (CVPR), pages 580–587, 2014
[43] Fawcett, T. 2006: An introduction to ROC analysis. Pattern recognition letters, 27(8), 861-874.
[44] Wilks, D., 2011: Statistical Methods in the Atmospheric Sciences. Third Edition.
[45] Swets, J. A., 1988: Measuring the accuracy of diagnostic systems. Science, 240(4857):1285-1293
[46] Hosmer, D. W., and Lemeshow, S., 2000: Multiple logistic regression (31-46). John Wiley & Sons Inc.